\documentclass{ceurart}
\usepackage{listings}
\usepackage{algorithm}
\usepackage{algorithmic}
\usepackage{tikz}
\usepackage{pgfplots}
\pgfplotsset{compat=1.18}
\usetikzlibrary{positioning,arrows.meta,shapes.geometric,fit,calc}

\newcommand{\pmsd}[2]{#1{\scriptstyle\,\pm\,#2}}

\begin{document}

\copyrightyear{2026}
\copyrightclause{Copyright for this paper by its authors.
  Use permitted under Creative Commons License Attribution 4.0
  International (CC BY 4.0).}

\conference{SuRE'26: 1st Workshop on Sustainability and Resource-Efficiency of
  Artificial Intelligence, co-located with IJCAI 2026,
  August 17, 2026, Bremen, Germany}

\title{ENAS: An Efficient Hardware-Aware Neural Architecture Search Framework
  for TinyML on Resource-Constrained Microcontrollers}

\author[1]{Mohd Moin Khan}[%
  email=moinkhanmohd@iisc.ac.in,
]
\author[1]{Naman Srivastava}[%
  email=snaman@iisc.ac.in,
]
\author[1]{Pandarasamy Arjunan}[%
  email=samy@iisc.ac.in,
]
\cormark[1]
\address[1]{Indian Institute of Science (IISc), Bengaluru 560012, India}

\cortext[1]{Corresponding author.}

\begin{abstract}
We present \textbf{ENAS}, a hardware-aware Neural Architecture Search (NAS)
framework that combines a static feasibility check, a cell-based search space
supporting standard, depthwise-separable, and bottleneck blocks with optional
skip connections, and a three-stage hybrid search strategy
(random $\rightarrow$ top-$K$ $\rightarrow$ mutation) with persistent
cross-run caching. Unlike many existing NAS frameworks that rely on GPU
acceleration, ENAS is designed to operate efficiently without requiring GPUs,
making it suitable for resource-constrained development environments. We
evaluate ENAS on two TinyML benchmarks, Visual Wake Words and Melanoma Cancer,
across eight microcontrollers with memory footprints ranging from 20\,KB to
1\,MB SRAM and nine input image resolutions. Our experimental results show that
ENAS achieves mean search-time speedups of $2.41{\times}$ and $1.70{\times}$ on
the Visual Wake Words and Melanoma Cancer datasets, respectively, while
maintaining competitive test accuracy compared with the recent NanoNAS
framework. A measured resource analysis further shows that ENAS-selected models
use substantially lower peak activation RAM, the binding constraint for
microcontroller deployment at matched accuracy. Additionally, ENAS achieves
$79.4\%$ test accuracy on an STM32H743-based microcontroller, outperforming the
greedy CPU-only baseline by $2.6$ percentage points. We release the ENAS
framework as open-source at:
\url{https://github.com/EdgeIntelligenceLab/ENAS}
\end{abstract}


\maketitle

\section{Introduction}

TinyML enables the deployment of deep learning models on microcontrollers
(MCUs) operating under kilobyte-scale SRAM and milliwatt-scale power budgets,
enabling always-on perception for battery-powered IoT devices. Designing
efficient TinyML models are largely driven by hardware-aware Neural Architecture
Search (NAS), where candidate models must satisfy strict deployment
constraints, including peak activation RAM, Flash storage for weights, and
multiply-accumulate (MACC) budgets that approximate inference latency and
energy consumption. The MACC budgets used in this paper follow the convention
established by NanoNAS~\cite{garavagno2024affordable}: they are application-driven latency targets rather than the hardware's raw capability, and are intentionally set below the peak device capability.

Recent TinyML NAS frameworks such as \textbf{MCUNet}~\cite{lin2020mcunet} and
\textbf{MicroNAS}~\cite{king2025micronas} achieve strong accuracy, but require
substantial GPU resources for search and training (e.g., MCUNet reports
approximately ${\sim}300$ GPU hours). Such requirements limit their
accessibility for many edge AI researchers, embedded developers, and small IoT
laboratories. More recent CPU-only NAS approaches, such as
\textbf{NanoNAS}~\cite{garavagno2024affordable}, a
hardware-aware NAS framework that has been shown to outperform several existing
TinyML NAS methods, improve accessibility by avoiding GPU dependence. However,
NanoNAS relies on a greedy search strategy over a relatively constrained search
space. It also incurs significant overhead from repeated TFLite conversion and
full model retraining during candidate evaluation. Furthermore, its search
space cannot effectively represent architectural primitives such as
depthwise-separable convolutions, bottleneck blocks, and skip connections,
which are known to be critical for efficient TinyML
deployments~\cite{tan2019efficientnet,sandler2018mobilenetv2}.

To address these limitations, this paper presents \textbf{ENAS}, a CPU-only
hardware-aware NAS framework for efficient TinyML model development targeting
computer vision applications on resource-constrained microcontrollers. Built on
top of the NanoNAS framework, ENAS extends its hardware-aware search pipeline
with lightweight analytical feasibility estimation, a richer cell-based search
space, and a more efficient hybrid search strategy. ENAS replaces NanoNAS's
per-candidate \emph{measured} feasibility check (which converts each candidate
to TFLite-Micro and measures RAM/Flash before deciding feasibility) with a
lightweight static \emph{analytical} estimator for RAM, Flash, and MACC
constraints (Eq.~\ref{eq:rmin}), eliminating most TFLite conversions during the
search. It further introduces a flexible cell-based search space supporting
standard, depthwise-separable, and bottleneck blocks with optional skip
connections. Instead of greedy coordinate ascent, ENAS employs a three-stage
hybrid strategy consisting of random sampling, top-$K$ candidate selection, and
local mutation, together with persistent cross-run caching to reduce redundant
evaluations.

Unlike prior work that typically evaluates NAS performance on a limited set of
hardware platforms or input configurations, we conduct a comprehensive study
across two TinyML applications (Visual Wake Words and Melanoma Cancer
classification), eight MCU platforms, and nine input image resolutions. This
results in 636 fully trained and deployed models, enabling a detailed analysis
of deployment-level behavior across diverse hardware-resource regimes,
including feasibility boundaries, search-time scaling, and the resource footprint
of selected architectures, and the impact of richer search spaces on
constrained MCUs.

\paragraph{Contributions.}
\begin{enumerate}\itemsep0pt
  \item We propose \textbf{ENAS}, a CPU-only hardware-aware NAS framework built
  on top of NanoNAS, combining a flexible cell-based search space and a
  three-stage hybrid search algorithm for TinyML deployment.

  \item We design analytical RAM, Flash, and MACC static feasibility checks
  that eliminate expensive late-stage TFLite conversion during candidate
  evaluation, reducing per-candidate evaluation time from $5$--$10$ minutes to
  approximately ${\sim}30$ seconds.

  \item We conduct a large-scale evaluation across 8 MCU platforms, 9 input
  resolutions, and 2 TinyML datasets, resulting in 636 fully trained models
  with zero deployment pipeline failures, and report mean$\pm$standard-deviation
  statistics together with Wilcoxon significance tests throughout.

  \item ENAS achieves mean search-time speedups of $2.41{\times}$ on Visual
  Wake Words and $1.70{\times}$ on Melanoma Cancer compared with the greedy
  CPU-only NanoNAS baseline (both highly significant, $p<10^{-8}$), while
  maintaining competitive accuracy with only small mean accuracy reductions of
  $0.79$\,pp and $1.31$\,pp, respectively.

  \item Using measured (on-tool) RAM and Flash footprints, we show that
  ENAS-selected architectures use markedly lower peak activation RAM
  (${\sim}0.38{\times}$ at matched accuracy on both datasets), the resource
  that binds first on MCUs, at the cost of higher Flash and MACC, and we
  identify deployment regimes where richer search spaces provide accuracy
  benefits (up to $+10.6$\,pp on STM32H743) as well as scenarios where the
  simpler greedy baseline remains preferable.
\end{enumerate}

\section{Related Works}

\paragraph{Neural architecture search.}
The NAS literature spans reinforcement-learning
controllers~\cite{tan2019mnasnet}, differentiable methods with Gumbel-softmax
relaxations~\cite{wu2019fbnet,cai2018proxylessnas}, evolutionary search, and
weight-sharing supernets~\cite{cai2019once}. These methods deliver
state-of-the-art accuracy at mobile latency budgets but require hundreds to
tens of thousands of GPU hours. Path binarisation~\cite{cai2018proxylessnas}
and one-shot supernet training~\cite{cai2019once} reduce this cost but still
target mobile and server hardware rather than kilobyte-scale microcontrollers.
Zero-cost proxies~\cite{abdelfattah2021zero} eliminate training during search,
though their rank correlation with final accuracy degrades on sub-100\,KB
architectures.


\paragraph{NAS for Microcontrollers.}
Microcontroller-targeted NAS introduces rigid SRAM and Flash constraints that compress the search space and necessitate tight co-design with downstream inference runtimes. Within this domain, MCUNet~\cite{lin2020mcunet} couples a two-stage TinyNAS search with the customized TinyEngine runtime, achieving over $70\%$ ImageNet top-1 accuracy on commercial MCUs; subsequent extensions integrate patch-based inference~\cite{lin2110mcunet} and enable on-device training within a tight $256$\,KB budget~\cite{lin2022device}. Similarly, MicroNAS~\cite{king2025micronas} leverages differentiable NAS alongside latency lookup tables for time-series classification on MCUs. To minimize optimization overhead, zero-shot MCU variants~\cite{qiao2024micronas,qiao2025monas} achieve substantial search speedups via training-free proxies, while the recent ELASTIC~\cite{tran2026elastic} framework expands once-for-all supernet search to object detection on microcontrollers. 

Crucially, all of these frameworks depend on GPU compute or intensive supernet pre-training. To eliminate this bottleneck, NanoNAS~\cite{garavagno2024affordable} descended from ColabNAS~\cite{garavagno2024colabnas} employs a CPU-only, greedy coordinate-ascent search over filter count $k$ and depth $c$, while TinyTNAS~\cite{saha2024tinytnas} introduces a parallel CPU-only, time-bounded approach for time-series workloads. The proposed ENAS framework belongs to this sustainable, CPU-only family; however, unlike ELASTIC and related supernet methodologies, it operates entirely without GPU acceleration or pre-training overhead, and unlike NanoNAS, it navigates a significantly richer, cell-based search space.

\paragraph{Limitations of existing CPU-only NAS.}
Three limitations of the greedy CPU-only baseline emerge from measured
experiments. \\ \textbf{(L1) Search-time bottleneck:} for every candidate, NanoNAS
performs a \emph{measured} feasibility check: it converts the candidate to
TFLite-Micro and measures peak RAM/Flash only \emph{after} the candidate has
been trained, so the search phase consumes 60--76\,\% of total runtime
($9$--$155$\,min per configuration). ENAS \emph{replaces this measured check with an analytical one}
performed before training. \\ \textbf{(L2) Narrow search space:} the $(k, c)$
parameterisation excludes depthwise-separable convolution, bottlenecks, skip
connections, and stride control. These primitives are parameter- and
activation-efficient; depthwise-separable and bottleneck blocks reduce
per-layer multiply-accumulate cost relative to standard
convolution~\cite{howard2017mobilenets,sandler2018mobilenetv2}. \\ \textbf{(L3)
Greedy instability:} per-board accuracy varies by up to 13\,pp between runs on
the same configuration on STM32H743 at $96{\times}96$, indicating local optima
in coordinate ascent.

\begin{figure*}[t]
\centering
\pgfdeclarelayer{background}
\pgfsetlayers{background,main}
\begin{tikzpicture}[
  font=\footnotesize,
  box/.style={draw, rounded corners=2pt, minimum height=0.62cm,
              align=center, inner sep=3pt},
  phase/.style={box, fill=black!8, minimum width=2.05cm},
  cell/.style={box, fill=black!4, minimum width=1.5cm},
  arr/.style={-{Latex[length=1.4mm]}, thick}
]
\node[phase] (pf)  {Phase 0\\Feasibility\\\scriptsize Eq.~\ref{eq:rmin}};
\node[phase, right=0.45cm of pf] (s)
    {Phase 1\\3-stage Search\\\scriptsize 70 candidates};
\node[phase, right=0.45cm of s]  (tr)
    {Phase 2\\Full Training\\\scriptsize 100 $E_p$, cosine LR};
\node[phase, right=0.45cm of tr] (q)
    {Phase 3\\INT8 PTQ\\\scriptsize + evaluation};
\draw[arr] (pf)--(s);
\draw[arr] (s)--(tr);
\draw[arr] (tr)--(q);
\node[cell, below=0.55cm of s.south, xshift=-1.7cm] (r)
    {Random\\\scriptsize $N_{\text{rand}} {=} 30$};
\node[cell, right=0.3cm of r]  (tk)
    {Top-$K$\\\scriptsize $K{=}8$};
\node[cell, right=0.3cm of tk] (mu)
    {Mutate\\\scriptsize Mutate (40 total)};
\begin{pgfonlayer}{background}
  \node[draw, rounded corners=4pt, fill=black!2,
        inner sep=6pt, fit=(r)(tk)(mu)] (substeps) {};
\end{pgfonlayer}
\draw[arr] (r)--(tk);
\draw[arr] (tk)--(mu);
\draw[arr, dashed] (s.south) -- (substeps.north);
\node[box, fill=black!8, minimum width=2.3cm, right=1.6cm of q] (stem)
    {Stem: Conv$3{\times}3$, $k$};
\node[box, fill=black!4, minimum width=2.3cm, below=0.3cm of stem] (c1)
    {Cell$_1$ \scriptsize(6-tuple)};
\node[box, fill=black!4, minimum width=2.3cm, below=0.3cm of c1]  (c2)
    {Cell$_2 \ldots$ Cell$_N$};
\node[box, fill=black!8, minimum width=2.3cm, below=0.3cm of c2]  (head)
    {GAP $\to$ Drop $\to$ Dense};
\draw[arr] (stem)--(c1);
\draw[arr] (c1)--(c2);
\draw[arr] (c2)--(head);
\node[above=0.1cm of stem, font=\scriptsize\itshape]
    {Architecture template};
\end{tikzpicture}
\caption{ENAS pipeline (left) and cell-based architecture template (right).
Phase~0 eliminates infeasible candidates analytically. Phase~1 evaluates 70
candidates (30 random + 40 mutants (5 per top-$K$ survivor)) using short proxy
training. Each cell is independently configured across six dimensions: block
type, kernel size, stride, skip connection, activation, and expansion ratio.}
\label{fig:pipeline}
\end{figure*}
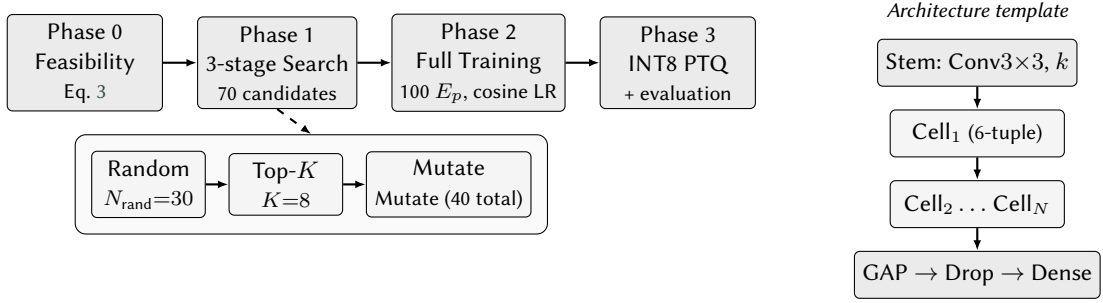

\section{The ENAS Framework}

Hardware-aware Neural Architecture Search (HW-NAS) for TinyML can be formulated
as a constrained optimisation problem. Given a search space of candidate
architectures $\mathcal{S}$, the goal is to identify an architecture
$\mathcal{A}^{\star} \in \mathcal{S}$ that maximises deployment accuracy while
satisfying the hardware constraints of a target microcontroller:
\begin{equation}
\mathcal{A}^{\star} = \arg\max_{\mathcal{A} \in \mathcal{S}}
\mathrm{Acc}(\mathcal{A})
\label{eq:objective}
\end{equation}
subject to:
\begin{equation}
R(\mathcal{A}) \le R_{\max}, \quad
F(\mathcal{A}) \le F_{\max}, \quad
M(\mathcal{A}) \le M_{\max},
\label{eq:constraints}
\end{equation}
where $R(\mathcal{A})$, $F(\mathcal{A})$, and $M(\mathcal{A})$ denote the peak
SRAM usage, Flash footprint, and multiply-accumulate (MACC) operations of
architecture $\mathcal{A}$, and $(R_{\max}, F_{\max}, M_{\max})$ are the
hardware budgets.

NanoNAS adopts a CPU-only greedy search and evaluates each candidate through
repeated TFLite conversion, proxy training, and a measured hardware feasibility
validation. While this reduces dependence on GPU clusters, the search remains
expensive due to repeated post-hoc measurement and retraining, and its
restricted $(k,c)$ space cannot capture efficient primitives such as
depthwise-separable convolutions, bottleneck layers, and skip connections.

To address these limitations, ENAS extends NanoNAS with three improvements:
\textbf{(1)} static analytical feasibility estimation to eliminate unnecessary
TFLite conversion, \textbf{(2)} a richer cell-based search space, and
\textbf{(3)} a three-stage hybrid search strategy with persistent cross-run
caching. ENAS is a four-phase pipeline (Fig.~\ref{fig:pipeline}): \textbf{(0)}
static feasibility check, \textbf{(1)} cell-based search, \textbf{(2)} full
training, and \textbf{(3)} INT8 post-training quantisation (PTQ) and evaluation.
The framework is designed for efficient operation on a single CPU node without
GPU acceleration.

\subsection{Static Feasibility Check}

Given hardware constraints $(R_{\max}, F_{\max}, M_{\max})$ and input shape
$(H, W, C_{\text{in}})$, ENAS computes a lower bound on peak activation memory:
\begin{equation}
R_{\min} = H \cdot W \cdot 4 + H \cdot W \cdot C_{\text{in}}
\quad \text{bytes}.
\label{eq:rmin}
\end{equation}
Equation~\eqref{eq:rmin} is \emph{not} a per-layer estimate: it is an
input-dependent lower bound on the network's peak activation footprint,
combining the largest single activation tensor that any feasible network must
materialise (the $H\cdot W\cdot4$ term, a working buffer) with the input tensor
itself ($H\cdot W\cdot C_{\text{in}}$). Because every architecture in
$\mathcal{S}$ must at least hold the input and one activation tensor, any
configuration with $R_{\min} > R_{\max}$ is provably infeasible and is rejected
before training. The check is therefore a necessary (not sufficient) condition
and assumes no relation such as $C_{\text{in}}=C_{\text{out}}$ in any layer;
intermediate-layer feasibility is verified analytically per candidate during
search (Algorithm~\ref{alg:enas}, line~4).

\begin{figure}[t]
\centering
\includegraphics[width=0.8\columnwidth]{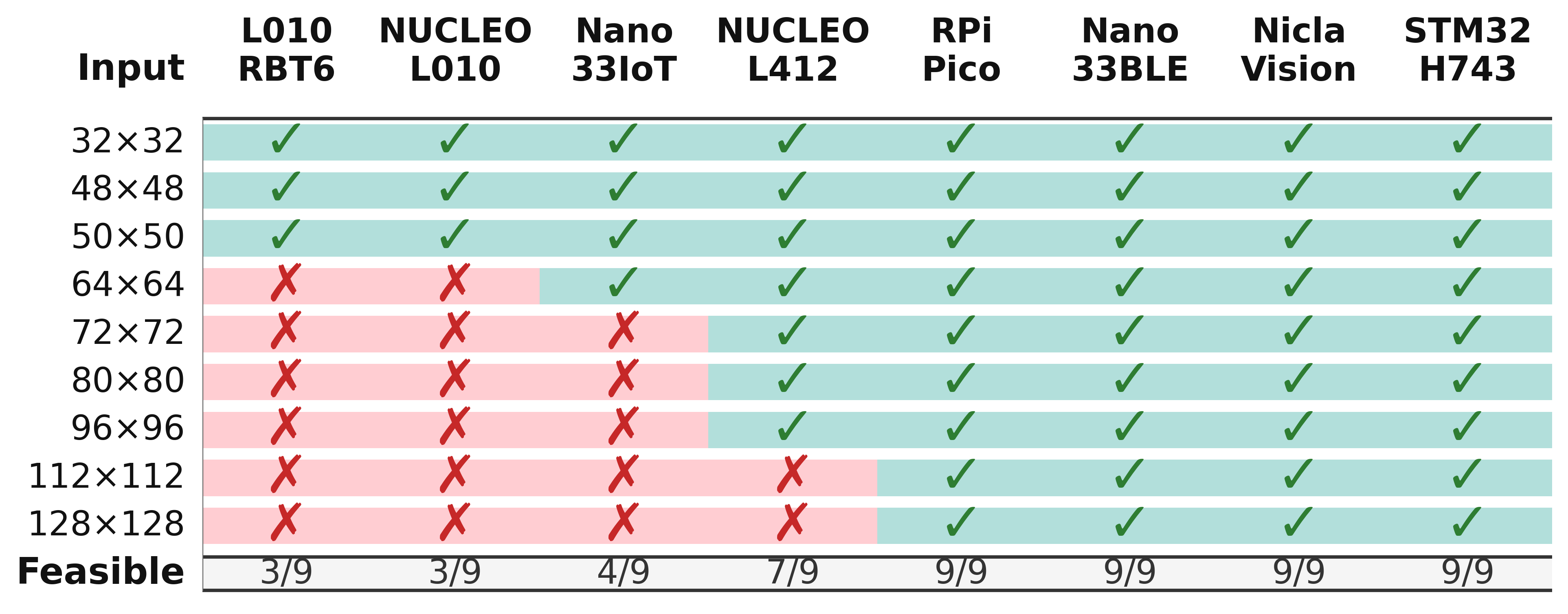}
\caption{Analytical activation-memory boundary. Peak RAM scales with
$H \cdot W \cdot k \cdot 4$ bytes, accurately predicting the infeasible regions
later detected by NanoNAS during its measured TFLite deployment check.}
\label{fig:ram-boundary}
\end{figure}

Across the 72 hardware-resolution configurations evaluated per dataset, the analytical check identified all infeasible cases in advance. As shown in Fig.~\ref{fig:ram-boundary}, infeasibility follows the predictable boundary
$R_{\min} \propto H \cdot W \cdot k \cdot 4$ bytes, matching the exact points where NanoNAS later fails during its measured post-hoc TFLite deployment check.

\subsection{Cell-Based Search Space}

Each architecture is represented as
$\mathcal{A} = (k, [\mathbf{c}_1, \ldots, \mathbf{c}_n])$,
where $k \in \{1,2,\ldots,16\}$ denotes the number of stem convolution filters
and $n \in \{1,\ldots,6\}$ the number of cells. Each cell $\mathbf{c}_i$ is
parameterised as $\mathbf{c}_i = (b, \kappa, s, g, a, e)$, where
$b \in \{$standard, depthwise-separable, bottleneck$\}$ is the block type,
$\kappa \in \{3,5\}$ the kernel size, $s \in \{1,2\}$ the stride,
$g \in \{$true, false$\}$ the skip-connection flag,
$a \in \{$\texttt{relu}, \texttt{relu6}$\}$ the activation, and
$e \in \{1,2\}$ the expansion ratio. During Stage~1, random sampling draws the
stem width from the constrained range $k\le 12$ to bias exploration toward
compact stems on tight budgets; Stage~3 mutation samples $k$ from the full
$\{1,\ldots,16\}$ range, so the realised search space spans
$k\in\{1,\ldots,16\}$ (selected values in our experiments range from $1$ to
$16$). Per-cell channel counts are \emph{not} searched independently: they are
derived deterministically from the stem width $k$ via a fixed
expansion schedule $n_{\text{out}}=\lceil n\cdot m_i\rceil$ with a decaying
multiplier $m_i$, mirroring NanoNAS's channel-doubling rule. This keeps the
search space tractable ($\sim$$10^{4}$ structural configurations after fixing
the channel schedule) while letting the cell tuple control the qualitatively
important primitives. The cell options are intentionally restricted to the
primitives with the highest accuracy-per-byte return on MCUs (small kernels,
binary stride, low expansion); broadening them is straightforward but enlarges
the search budget (Section~\ref{sec:limitations}). Skip connections improve
gradient flow when tensor dimensions match, while \texttt{relu6} improves
compatibility with INT8 PTQ due to its bounded activation range.

\subsection{Three-Stage Hybrid Search}
\label{sec:algo}

Algorithm~\ref{alg:enas} presents the overall search strategy for ENAS. It approximates the optimal architecture through a
three-stage hybrid strategy. ENAS first evaluates $N_{\text{rand}}{=}30$
randomly sampled architectures, selects the top $K{=}8$ candidates, and
generates $N_{\text{mut}}{=}5$ mutations per survivor, producing 40 mutated
candidates in total. The budget $(N_{\text{rand}}, K, N_{\text{mut}})=(30,8,5)$
yields a 70-candidate pass that balances feasible-region coverage and search
cost on resource-constrained MCUs: $N_{\text{rand}}=30$ ensures sufficient
feasible exploration, $K=8$ focuses mutation on high-performing candidates, and
$N_{\text{mut}}=5$ provides near-complete coverage of the single-edit
neighbourhood of the 6-dimensional cell space. To avoid redundant evaluations,
ENAS maintains a persistent cross-run cache of previously evaluated
architectures for the same hardware and input configuration.

Each candidate is scored using:
\begin{equation}
\begin{aligned}
s &= w_{\text{acc}}\, v_{\text{acc}}
   + w_{\text{eff}}\Big(1 - \sqrt[3]{\tfrac{r}{R_{\max}}
     \tfrac{f}{F_{\max}}\tfrac{m}{M_{\max}}}\Big)
   + w_{\text{hr}}\, h,
\end{aligned}
\label{eq:score}
\end{equation}
where
$h = \mathbf{1}\!\left[\max\!\left(\tfrac{r}{R_{\max}},\tfrac{f}{F_{\max}},
\tfrac{m}{M_{\max}}\right)\le 0.8\right]$.
The three terms encode complementary objectives.
$v_{\text{acc}}$ is the proxy validation accuracy after $E_p{=}3$ epochs on
30\,\% of the training data, and $w_{\text{acc}}$ weights it.
$w_{\text{eff}}$ weights an \emph{efficiency} term: one minus the geometric
mean of the three normalised resource ratios, which rewards architectures that
leave headroom on all three budgets simultaneously (the geometric mean prevents
a single slack constraint from masking a tight one). $w_{\text{hr}}$ weights a
binary \emph{headroom} indicator $h$, which fires only when \emph{every}
resource ratio is $\le 0.8$; it is a discrete bonus, deliberately separate from
the continuous efficiency term, that protects a margin for INT8 PTQ overhead and
fine-tuning that the smooth efficiency term alone does not guarantee at the
$0.8$ boundary. We set
$(w_{\text{acc}}, w_{\text{eff}}, w_{\text{hr}}) = (0.80, 0.15, 0.05)$.

Stage~2 performs a two-phase ranking. First, candidates are filtered by proxy
validation accuracy alone; the remaining top 50\,\% are then ranked using the full scoring function in Eq.~\ref{eq:score}, and the top-$K$ architectures are retained for mutation. When a previously feasible architecture exists for the same hardware platform, ENAS may inject it as a \emph{warm seed} (refer to Section~\ref{sec:protocol}) into the
Stage~1 candidate pool.

\begin{algorithm}[t]
\caption{ENAS Hybrid Search}
\label{alg:enas}
\textbf{Input}: hardware constraints $(R_{\max}, F_{\max}, M_{\max})$,
input shape $\mathbf{x}$, proxy epochs $E_p$, scoring weights
$(w_{\text{acc}}, w_{\text{eff}}, w_{\text{hr}})$, search budget
$(N_{\text{rand}}, K, N_{\text{mut}})$, optional warm-seed architecture
$\mathcal{A}_0$.\\
\textbf{Output}: best architecture $\mathcal{A}^{\star}$
\begin{algorithmic}[1]
\STATE Compute $R_{\min}$ via Eq.~\eqref{eq:rmin}; if $R_{\min} > R_{\max}$,
       \textbf{return infeasible}.
\STATE Load persistent cache $\mathcal{C}$.
\STATE \emph{Stage 1 (Random).} Sample $N_{\text{rand}}$ random architectures;
       if $\mathcal{A}_0$ is available, include it in the pool.
\STATE In parallel, for each $\mathcal{A}$: query $\mathcal{C}$; else estimate
       $(r,f,m)$ analytically. If any constraint is violated, mark infeasible;
       otherwise proxy-train $E_p$ epochs on 30\,\% data and score by
       Eq.~\eqref{eq:score}.
\STATE \emph{Stage 2 (Top-$K$).} Apply the accuracy-first filter, then the
       full score of Eq.~\eqref{eq:score}, and retain the top $K$.
\STATE \emph{Stage 3 (Mutation).} Generate $N_{\text{mut}}$ mutated
       architectures per survivor and evaluate them in parallel using
       Eq.~\eqref{eq:score}.
\STATE \textbf{return} $\mathcal{A}^{\star} = \arg\max s(\mathcal{A})$ over
       Stages 1 and 3.
\end{algorithmic}
\end{algorithm}

\subsection{Calibration of Proxy Training and Scoring}
\label{sec:calibration}

The effectiveness of ENAS depends on the proxy training budget $E_p$ and the
weight $w_{\text{acc}}$. An initial configuration using $E_p{=}1$ and
$w_{\text{acc}}{=}0.65$ produced a Spearman rank correlation of $\rho{=}0.18$
($n{=}30$, not significant) between proxy validation accuracy and final TFLite
INT8 test accuracy, too low for reliable ranking. The richer cell-based space
introduces skip-connection paths and deeper architectures whose optimisation
dynamics require more than a single proxy epoch to become informative.
Increasing $E_p$ to 3 improves the rank correlation to $\rho{=}0.71$
($p{<}0.001$). Increasing $w_{\text{acc}}$ to $0.80$ further prevents the
efficiency term from dominating selection when proxy accuracies fall within a
narrow range. All ENAS results use the calibrated setting
$(E_p{=}3, w_{\text{acc}}{=}0.80)$ unless otherwise specified.

\section{Experimental Setup}

\subsection{Datasets}
We evaluate on two TinyML benchmarks chosen for their distinct difficulty
regimes: \textbf{Visual Wake Words}~\cite{chowdhery2019visual}, binary
person-presence on natural images (a vision benchmark of moderate difficulty,
${\sim}73\%$ baseline accuracy), and \textbf{Melanoma Cancer}, a
medical-imaging benchmark with binary classification of dermoscopic images and
stronger class signal (${\sim}89\%$ baseline accuracy). Together they test
whether ENAS generalises across signal regimes.

\subsection{Hardware Platforms}
The evaluation considers eight distinct MCU platforms (Table~\ref{tab:hw}) spanning three resource tiers: ultra-constrained (with $20$\,KB of SRAM), mid-tier, and high-capacity (up to $1$\,MB of SRAM). Across these platforms, the target hardware configuration budgets range from $0.75$\,M to $15.00$\,M MACC operations.
\begin{table}[t]
\centering\small
\caption{Target MCU platforms.}
\begin{tabular}{lrrr}
\toprule
\textbf{Platform} & \textbf{RAM} & \textbf{Flash} & \textbf{MACC}\\
\midrule
STM32L010RBT6        & 20\,KB  & 128\,KB & 0.75\,M\\
NUCLEO-L010RB        & 20\,KB  & 64\,KB  & 0.75\,M\\
Arduino Nano33IoT    & 32\,KB  & 256\,KB & 1.20\,M\\
NUCLEO-L412KB        & 64\,KB  & 128\,KB & 3.20\,M\\
Raspberry Pi Pico    & 264\,KB & 2\,MB   & 3.00\,M\\
Arduino Nano33BLE    & 256\,KB & 1\,MB   & 4.00\,M\\
Arduino Nicla Vision & 1\,MB   & 2\,MB   & 8.00\,M\\
STM32H743ZI          & 1\,MB   & 2\,MB   & 15.0\,M\\
\bottomrule
\end{tabular}
\label{tab:hw}
\end{table}

\subsection{Protocol}
\label{sec:protocol}
We evaluate square input resolutions $P_X \times P_X$, with
$P_X \in \{32, 48, 50, 64, 72, 80, 96, 112, 128\}$. Across the resulting
$8 \times 9 = 72$ (hardware, resolution) configurations, 53 are feasible under
Eq.~\eqref{eq:rmin} and 19 are correctly rejected. We run each feasible cell
three times per method per dataset, totalling
$53 \times 3 \times 2 \times 2 = 636$ fully trained models. Full training uses
100 epochs with cosine LR decay ($10^{-2}{\to}10^{-6}$), batch size 128, and
\texttt{EarlyStopping} (patience 15). INT8 PTQ uses 150 representative samples;
peak RAM and Flash of the resulting model are measured with
STMicroelectronics' \texttt{stm32tflm} tool, identical to NanoNAS, so all
measured resource numbers are directly comparable.

\paragraph{Warm seeding.}
A \emph{warm seed} is an architecture from a prior, already-completed search on
the \emph{same} hardware/resolution encoded as a standard-convolution cell
sequence with the NanoNAS-best $(k,c)$, that is prepended to the Stage~1
candidate pool. It guarantees a non-empty feasible pool on tight RAM budgets
where random sampling may produce no feasible cell, and anchors top-$K$
selection with a known-good accuracy floor. Unless stated otherwise, the broad
cross-resolution sweep (Tables~\ref{tab:agg}, \ref{tab:combined},
\ref{tab:perhw-vww}, \ref{tab:resource}) is run \emph{without} warm seeding; the
focused $50{\times}50$/$64{\times}64$ study (Table~\ref{tab:focus}) uses warm
seeding, which is why its selected architectures differ. We compare ENAS
against NanoNAS~\cite{garavagno2024affordable} on identical hardware budgets and
report an intermediate ablation (\emph{ENAS-strategy}) that retains the 2-D
search space with parallel random search to isolate strategy from search space.

\section{Results}

\subsection{Aggregate Trade-off}

\begin{table}[t]
\centering\small
\caption{Aggregate results across all 53 feasible (hardware, resolution) cells
per dataset. Each cell statistic is the mean over 3 runs; aggregate values are
mean$\pm$std over the 53 cell means. $p$-values are from two-sided Wilcoxon signed-rank tests comparing ENAS and NanoNAS for each dataset.}
\label{tab:agg}
\begin{tabular}{lcccc}
\toprule
\textbf{Metric} & \multicolumn{2}{c}{\textbf{VWW}} & \multicolumn{2}{c}{\textbf{Cancer}} \\
\cmidrule(lr){2-3}\cmidrule(lr){4-5}
 & \textbf{NanoNAS} & \textbf{ENAS} & \textbf{NanoNAS} & \textbf{ENAS} \\
\midrule
Mean acc.\ (\%)
  & $\pmsd{72.68}{2.58}$ & $\pmsd{71.89}{3.33}$
  & $\pmsd{89.44}{1.26}$ & $\pmsd{88.13}{1.52}$ \\
Median acc.\ (\%) & 72.71 & 72.24 & 89.77 & 88.23 \\
Mean search (min)
  & $\pmsd{102.2}{44.6}$ & $\pmsd{42.4}{21.2}$
  & $\pmsd{10.3}{3.8}$   & $\pmsd{6.1}{2.6}$ \\
  \midrule
\textbf{Speedup}  & \multicolumn{2}{c}{\textbf{2.41$\times$}}
                  & \multicolumn{2}{c}{\textbf{1.70$\times$}} \\
Acc.\ $p$-value   & \multicolumn{2}{c}{$0.118$ (n.s.)}
                  & \multicolumn{2}{c}{$3.4\times10^{-6}$} \\
Search $p$-value  & \multicolumn{2}{c}{$4.7\times10^{-10}$}
                  & \multicolumn{2}{c}{$5.9\times10^{-9}$} \\
\bottomrule
\end{tabular}
\end{table}

Table~\ref{tab:agg} summarizes the aggregate performance metrics. ENAS achieves a mean search-time speedup of $2.41{\times}$ on VWW and $1.70{\times}$ on Cancer, accompanied by minor average accuracy reductions of $0.79$\,pp and $1.31$\,pp, respectively. Statistical analysis reveals a clear distinction between these two outcomes: the search-time reduction is highly significant across both datasets ($p < 10^{-8}$), whereas the accuracy difference is not statistically significant on VWW ($p = 0.118$) and remains small on Cancer despite reaching significance.
Standard deviations are slightly higher under ENAS, a consequence of the larger search space exposing more architectural variation, but median accuracy gaps remain tight (0.47\,pp and 1.54\,pp). Search time is the discriminative dimension: ENAS strictly dominates on per-configuration runtime.



\subsection{Resource Footprint of Selected Models}
\label{sec:resource}

\begin{table}[t]
\centering\small
\caption{Measured resource footprint of selected models, mean$\pm$std over the
53 feasible cells (RAM and Flash measured with \texttt{stm32tflm}; MACC
analytical). The final two rows give the ENAS/NanoNAS ratio over all cells and
restricted to matched-accuracy cells ($|\Delta\text{acc}|\le1$\,pp).}
\label{tab:resource}
\setlength{\tabcolsep}{4pt}
\begin{tabular}{llccc}
\toprule
\textbf{Dataset} & \textbf{Method} & \textbf{RAM (KB)} & \textbf{Flash (KB)} & \textbf{MACC (M)}\\
\midrule
\multirow{2}{*}{VWW}
 & NanoNAS & $\pmsd{40.4}{28.6}$ & $\pmsd{14.0}{4.6}$  & $\pmsd{1.77}{1.49}$\\
 & ENAS    & $\pmsd{12.1}{5.5}$  & $\pmsd{53.8}{37.8}$ & $\pmsd{2.56}{2.44}$\\
\midrule
\multirow{2}{*}{Cancer}
 & NanoNAS & $\pmsd{37.0}{27.3}$ & $\pmsd{9.4}{2.4}$   & $\pmsd{0.86}{0.86}$\\
 & ENAS    & $\pmsd{7.6}{4.1}$   & $\pmsd{50.7}{42.0}$ & $\pmsd{1.85}{1.72}$\\
\midrule
\multicolumn{2}{l}{Ratio (all, VWW / Cancer)}
 & 0.30 / 0.21 & 3.84 / 5.40 & 1.44 / 2.15\\
\multicolumn{2}{l}{Ratio (matched acc.)}
 & 0.38 / 0.38 & 2.85 / 5.73 & 1.39 / 3.94\\
\bottomrule
\end{tabular}
\end{table}

Table~\ref{tab:resource} evaluates whether richer architectural primitives yield superior efficiency in deployed models. The results reveal a nuanced trade-off that is highly advantageous for TinyML microcontrollers. Specifically, ENAS-selected architectures require significantly less peak activation RAM than NanoNAS: approximately $0.30{\times}$ on VWW and $0.21{\times}$ on Cancer across all cells, and a consistent $0.38{\times}$ on both datasets within the matched-accuracy subset. Conversely, ENAS models utilize more Flash ($2.85$--$5.73{\times}$ at matched accuracy) and higher MACC ($1.39$--$3.94{\times}$). 
This trade-off aligns precisely with real-world microcontroller constraints, where peak SRAM serves as the primary binding bottleneck; indeed, every infeasible cell encountered in this study was strictly RAM-infeasible, making the activation-memory bound in Eq.~\eqref{eq:rmin} the critical limiting factor.  In contrast, Flash and computational capacity are comparatively abundant on these platforms, as evidenced by the fact that all 636 selected models easily fit within the physical hardware budgets.
The depthwise-separable and bottleneck blocks, together with stride-based downsampling and skip connections, let ENAS produce architectures with smaller peak activation tensors at comparable accuracy. This demonstrates that the cell-based framework does not simply offer an alternative search space, but actively biases optimization toward RAM-efficient topologies. Finally, while the analytical Flash term in Eq.~\eqref{eq:score} serves as a loose lower bound for early filtering, the empirical \texttt{stm32tflm} measurements confirm that physical feasibility holds in practice because the conservative RAM estimate remains the definitive pre-flight constraint.

\begin{table}[t]
\centering\small
\caption{TFLite INT8 test accuracy (\%, mean$\pm$std over 3 runs) at the two
most commonly-targeted TinyML input sizes on VWW, using warm seeding (see
Section~\ref{sec:protocol}). ENAS surpasses NanoNAS on the mean at
$64{\times}64$ and on three of six platforms.}
\setlength{\tabcolsep}{4pt}
\begin{minipage}[t]{0.49\textwidth}
\centering
\textbf{Input $50{\times}50$}\\[2pt]
\begin{tabular}{lccr}
\toprule
\textbf{Hardware} & \textbf{NanoNAS} & \textbf{ENAS} & $\Delta$\\
\midrule
L010RBT6     & $\pmsd{72.7}{1.4}$ & $\pmsd{69.9}{3.1}$ & $-2.9$\\
NUCLEO-L010  & $\pmsd{72.0}{1.3}$ & $\pmsd{68.1}{3.8}$ & $-3.9$\\
Nano33IoT    & $\pmsd{71.8}{1.9}$ & $\pmsd{70.4}{1.2}$ & $-1.4$\\
NUCLEO-L412  & $\pmsd{72.0}{2.7}$ & $\pmsd{71.2}{2.3}$ & $-0.8$\\
RPi Pico     & $\pmsd{73.5}{1.1}$ & $\pmsd{72.0}{1.0}$ & $-1.5$\\
Nano33BLE    & $\pmsd{73.7}{0.2}$ & $\pmsd{73.0}{1.4}$ & $-0.7$\\
Nicla Vision & $\pmsd{74.6}{0.2}$ & $\pmsd{74.1}{3.1}$ & $-0.5$\\
STM32H743    & $\pmsd{70.6}{6.1}$ & $\pmsd{75.1}{2.5}$ & $+4.5$\\
\midrule
Mean         & 72.6 & 71.7 & $-0.9$\\
\bottomrule
\end{tabular}
\end{minipage}%
\hfill
\begin{minipage}[t]{0.49\textwidth}
\centering
\textbf{Input $64{\times}64$}\\[2pt]
\begin{tabular}{lccr}
\toprule
\textbf{Hardware} & \textbf{NanoNAS} & \textbf{ENAS} & $\Delta$\\
\midrule
Nano33IoT    & $\pmsd{71.2}{3.2}$ & $\pmsd{72.0}{1.1}$ & $+0.8$\\
NUCLEO-L412  & $\pmsd{74.0}{1.8}$ & $\pmsd{72.0}{1.7}$ & $-2.0$\\
RPi Pico     & $\pmsd{72.9}{3.4}$ & $\pmsd{74.4}{1.3}$ & $+1.4$\\
Nano33BLE    & $\pmsd{71.2}{4.1}$ & $\pmsd{73.8}{1.8}$ & $+2.7$\\
Nicla Vision & $\pmsd{73.2}{2.4}$ & $\pmsd{75.9}{2.0}$ & $+2.6$\\
STM32H743    & $\pmsd{74.7}{1.6}$ & $\pmsd{74.7}{3.1}$ & $+0.0$\\
\midrule
Mean         & 73.0 & 73.7 & $+0.7$\\
\bottomrule
\end{tabular}
\end{minipage}
\label{tab:focus}
\end{table}

\subsection{Focused Study at 50$\times$50 and 64$\times$64 (VWW)}
Across the two most frequently targeted input dimensions (Table~\ref{tab:focus}), ENAS achieves competitive mean accuracy, yielding a minor deficit of $-0.90$\,pp at a $50{\times}50$ resolution and a net gain of $+0.70$\,pp at $64{\times}64$. Notably, at the $64{\times}64$ input resolution, ENAS outperforms the baseline alternative on four of the six feasible hardware platforms, which includes an accuracy improvement of $+2.60$\,pp on the Nicla Vision and $+2.70$\,pp on the Nano33BLE.

\subsection{Cross-Resolution Sweep}

\begin{table*}[t]
\centering\small
\caption{Per-resolution accuracy (\%, mean$\pm$std over the $N$ feasible cells)
and search time (min) across all feasible hardware platforms.
$\Delta = \text{ENAS}-\text{NanoNAS}$ in percentage points; speedup
$= \text{NanoNAS}/\text{ENAS}$. The mean speedups ($2.41{\times}$ on VWW,
$1.70{\times}$ on Cancer) hold uniformly across resolutions, while accuracy
varies by tier and input size.}
\label{tab:combined}
\setlength{\tabcolsep}{4pt}
\resizebox{\textwidth}{!}{%
\begin{tabular}{r r ccr cc c ccr cc c}
\toprule
 & & \multicolumn{6}{c}{\textbf{Visual Wake Words}} &
     \multicolumn{6}{c}{\textbf{Melanoma Cancer}} \\
\cmidrule(lr){3-8}\cmidrule(lr){9-14}
 & & \multicolumn{3}{c}{\emph{Accuracy (\%)}} &
     \multicolumn{3}{c}{\emph{Search (min)}} &
     \multicolumn{3}{c}{\emph{Accuracy (\%)}} &
     \multicolumn{3}{c}{\emph{Search (min)}} \\
\cmidrule(lr){3-5}\cmidrule(lr){6-8}\cmidrule(lr){9-11}\cmidrule(lr){12-14}
\textbf{Input} & $N$ &
 \textbf{NanoNAS} & \textbf{ENAS} & $\Delta$ &
 \textbf{Nano} & \textbf{ENAS} & \textbf{Sp.} &
 \textbf{NanoNAS} & \textbf{ENAS} & $\Delta$ &
 \textbf{Nano} & \textbf{ENAS} & \textbf{Sp.} \\
\midrule
$32^2$  & 8 & $\pmsd{69.6}{3.1}$ & $\pmsd{71.6}{1.9}$ & $+2.05$ & 58 & 39 & 1.50 & $\pmsd{89.9}{1.1}$ & $\pmsd{89.2}{1.3}$ & $-0.69$ & 9  & 6 & 1.49\\
$48^2$  & 8 & $\pmsd{71.4}{1.2}$ & $\pmsd{72.1}{4.3}$ & $+0.66$ & 82 & 37 & 2.25 & $\pmsd{90.0}{1.0}$ & $\pmsd{88.4}{1.5}$ & $-1.57$ & 10 & 5 & 1.82\\
$50^2$  & 8 & $\pmsd{72.6}{1.3}$ & $\pmsd{70.9}{3.8}$ & $-1.77$ & 103& 37 & 2.80 & $\pmsd{89.6}{1.0}$ & $\pmsd{88.4}{1.0}$ & $-1.22$ & 10 & 5 & 1.90\\
$64^2$  & 6 & $\pmsd{72.9}{1.4}$ & $\pmsd{72.6}{3.4}$ & $-0.24$ & 108& 41 & 2.62 & $\pmsd{89.1}{1.8}$ & $\pmsd{88.5}{1.1}$ & $-0.58$ & 10 & 6 & 1.64\\
$72^2$  & 5 & $\pmsd{74.0}{0.8}$ & $\pmsd{72.7}{3.2}$ & $-1.29$ & 120& 47 & 2.53 & $\pmsd{88.4}{1.7}$ & $\pmsd{88.3}{1.4}$ & $-0.12$ & 8  & 6 & 1.29\\
$80^2$  & 5 & $\pmsd{72.4}{2.8}$ & $\pmsd{72.9}{4.3}$ & $+0.52$ & 91 & 47 & 1.92 & $\pmsd{89.4}{1.1}$ & $\pmsd{88.5}{0.6}$ & $-0.83$ & 12 & 7 & 1.73\\
$96^2$  & 5 & $\pmsd{75.7}{1.2}$ & $\pmsd{72.1}{3.4}$ & $-3.62$ & 150& 48 & 3.11 & $\pmsd{89.9}{0.7}$ & $\pmsd{86.4}{0.1}$ & $-3.51$ & 13 & 7 & 1.97\\
$112^2$ & 4 & $\pmsd{75.0}{1.6}$ & $\pmsd{70.7}{4.4}$ & $-4.27$ & 144& 51 & 2.84 & $\pmsd{89.3}{1.5}$ & $\pmsd{87.3}{2.0}$ & $-2.02$ & 13 & 7 & 1.87\\
$128^2$ & 4 & $\pmsd{73.8}{3.2}$ & $\pmsd{71.6}{2.5}$ & $-2.21$ & 111& 47 & 2.36 & $\pmsd{88.4}{1.0}$ & $\pmsd{86.5}{2.4}$ & $-1.93$ & 10 & 6 & 1.61\\
\midrule
\textbf{Mean} & 53 & $\pmsd{72.7}{2.6}$ & $\pmsd{71.9}{3.3}$ & $-0.79$ & 102 & 42 & \textbf{2.41} & $\pmsd{89.4}{1.3}$ & $\pmsd{88.1}{1.5}$ & $-1.31$ & 10 & 6 & \textbf{1.70}\\
\bottomrule
\end{tabular}}
\end{table*}

Table~\ref{tab:combined} consolidates the per-resolution accuracy and search-time performance metrics across both datasets. Two distinct behavioral patterns emerge from these data. First, small-input regimes systematically favor ENAS; for instance, at a $32{\times}32$ resolution on VWW, ENAS yields an accuracy improvement of $+2.05$\,percentage points (pp). In this low-resolution regime, the richer architectural primitives particularly depthwise-separable cells utilizing a stride of 2 effectively compensate for constrained spatial information. Second, large-input regimes conversely favor NanoNAS, with ENAS exhibiting an accuracy deficit of $2.00\text{--}4.30$\,pp across the $96{\times}96\text{--}112{\times}112$ resolution range on both datasets. 

This performance degradation stems from a proxy-ranking limitation rather than an inherent deficiency in the search space. At larger input resolutions, the expanded feasible space accommodates numerous candidate architectures whose 3-epoch proxy validation accuracies are statistically indistinguishable, yet their fully converged final accuracies diverge substantially. 
Consequently, the brief proxy evaluation phase fails to reliably rank the upper tail of the architecture distribution, a phenomenon directly reflected in the elevated standard deviation of ENAS at these resolutions. The same
mechanism explains the consistent $50{\times}50$ shortfall: $50{\times}50$ sits
just above the feasibility boundary for the two 20\,KB devices, leaving a thin
feasible slice that the greedy baseline exhausts systematically but that
ENAS's 30-sample Stage~1 covers only sparsely. Crucially, the search-time
speedup is robust across the full range: every resolution gives a
$\ge 1.29{\times}$ speedup, and the speedup grows with resolution on VWW
(peaking at $3.11{\times}$ at $96{\times}96$) because NanoNAS's TFLite
conversion cost scales with model size while ENAS's analytical pre-flight
remains constant.

\subsection{Pareto and Per-Hardware Decomposition}

\begin{figure}[t]
\centering
\includegraphics[width=\columnwidth,height=8cm,keepaspectratio]{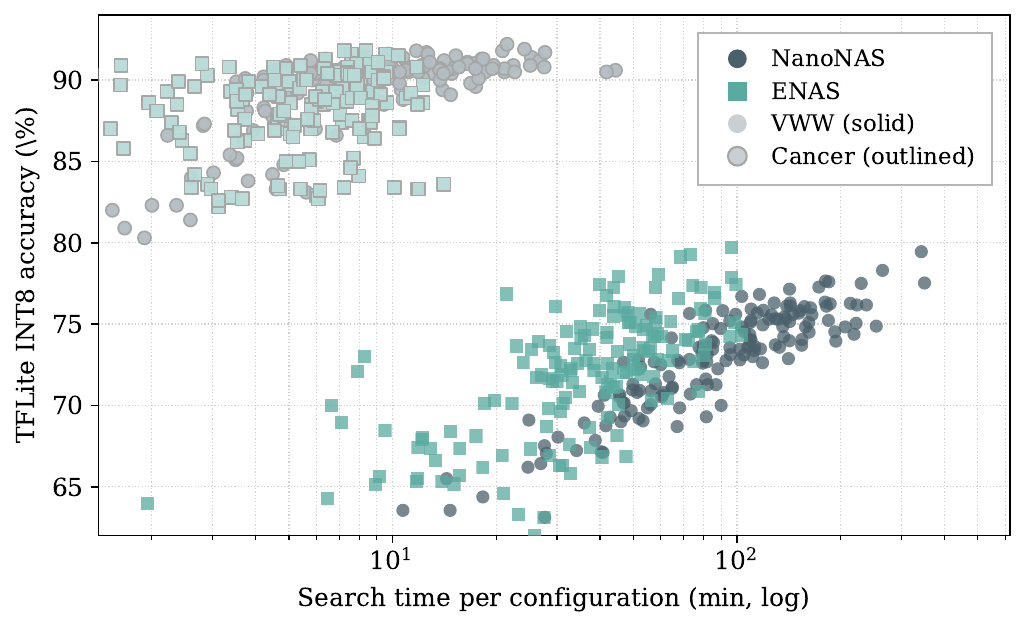}
\caption{Accuracy versus search-time Pareto across all 212 cells. ENAS (teal
squares) cluster at lower search times; NanoNAS (dark circles) spans a wider
accuracy range. Cross-dataset stratification is visible: Cancer at top
($\sim$89\%), VWW at bottom ($\sim$73\%).}
\label{fig:pareto}
\end{figure}

Fig.~\ref{fig:pareto} places all 212 (method, dataset, hardware, input) cells on
the accuracy--search-time plane; ENAS forms a tight low-cost cluster while
NanoNAS spans a wider range that extends to higher accuracy at higher search
time.

\begin{table}[t]
\centering\small
\caption{Per-hardware best TFLite accuracy on VWW (mean$\pm$std over 3 runs at
the best-accuracy resolution). ``Px'' is that resolution; ``Speed'' is the ENAS
search-time advantage at the ENAS-best configuration.}
\label{tab:perhw-vww}
\setlength{\tabcolsep}{3pt}
\begin{tabular}{lcccccc}
\toprule
\textbf{HW} & \textbf{RAM} & \multicolumn{2}{c}{\textbf{NanoNAS}} & \multicolumn{2}{c}{\textbf{ENAS}} & \textbf{Speed} \\
\cmidrule(lr){3-4}\cmidrule(lr){5-6}
 & (KB) & acc & Px & acc & Px & up \\
\midrule
L010RBT6 & 20  & $\pmsd{72.7}{1.4}$ & $50$ & $\pmsd{70.7}{3.4}$ & $32$ & 2.44$\times$\\
L010RB   & 20  & $\pmsd{72.0}{1.3}$ & $50$ & $\pmsd{68.3}{3.1}$ & $32$ & 2.79$\times$\\
IoT      & 32  & $\pmsd{72.7}{1.4}$ & $32$ & $\pmsd{70.4}{3.6}$ & $32$ & 2.84$\times$\\
L412     & 64  & $\pmsd{75.5}{0.6}$ & $80$ & $\pmsd{74.0}{3.5}$ & $48$ & 1.65$\times$\\
Pico     & 264 & $\pmsd{75.9}{0.5}$ & $96$ & $\pmsd{74.3}{1.9}$ & $50$ & 3.48$\times$\\
BLE      & 256 & $\pmsd{76.1}{1.1}$ & $128$& $\pmsd{75.6}{3.1}$ & $48$ & 3.09$\times$\\
Nicla    & 1024& $\pmsd{77.1}{0.7}$ & $128$& $\pmsd{75.5}{1.5}$ & $50$ & 3.02$\times$\\
H743     & 1024& $\pmsd{76.8}{1.4}$ & $96$ & $\pmsd{79.4}{0.3}$ & $80$ & 1.10$\times$\\
\bottomrule
\end{tabular}
\end{table}

\begin{figure*}[t]
\centering
\includegraphics[width=\columnwidth]{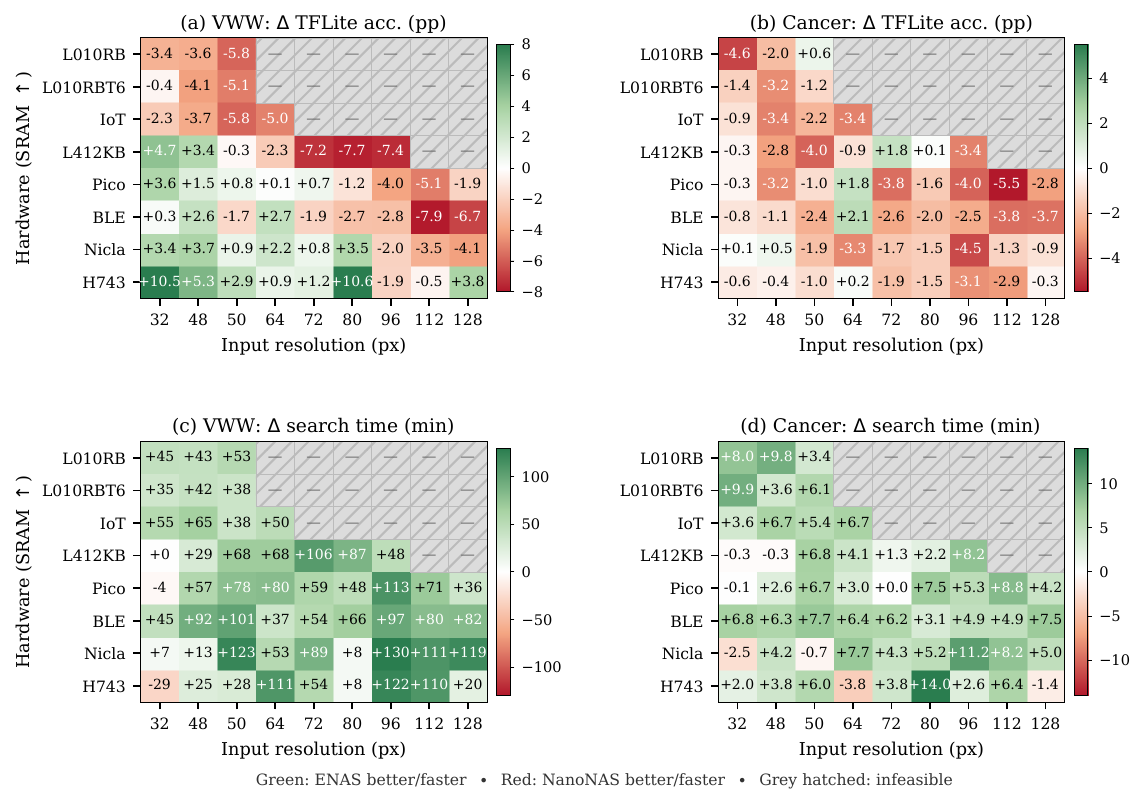}
\caption{Per-cell deltas (ENAS vs.\ NanoNAS) across both datasets. Top row:
$\Delta$ TFLite accuracy in percentage points (green $=$ ENAS better). Bottom
row: $\Delta$ search time in minutes (green $=$ ENAS faster). The H743 row in
panel~(a) shows the strongest ENAS dominance ($+10.5$\,pp at $32{\times}32$,
$+10.6$\,pp at $80{\times}80$); the search-time panels are almost uniformly
green. Grey hatched cells are RAM-infeasible.}
\label{fig:delta}
\end{figure*}

Table~\ref{tab:perhw-vww} and Fig.~\ref{fig:delta} decompose accuracy and
search time across hardware. NanoNAS wins seven of eight per-hardware bests on
VWW at $1.65$--$3.48{\times}$ longer search time, with the largest advantages on
mid-tier platforms (Pico, BLE, Nicla). ENAS wins decisively on STM32H743
($+2.6$\,pp at near-equal search time): the cell-based search space exposes a
deep architecture with a $k{=}13$ stem reaching $79.4\%$ TFLite accuracy at
$80{\times}80$ (std $0.3$\,pp), the single best result in our study, and this
advantage holds across multiple H743 resolutions ($+10.5$\,pp at $32{\times}32$,
$+10.6$\,pp at $80{\times}80$). Because the search space admits stems up to
$k=16$, these wider high-capacity-MCU architectures lie entirely outside the
NanoNAS $(k,c)$ space.

\paragraph{Strategy vs.\ search space.}
To isolate the search strategy from the search space, we evaluated an
intermediate variant (\emph{ENAS-strategy}) retaining the original 2-D $(k, c)$
space but using parallel random search with the analytical pre-flight check.
This variant captures the search-time speedup but none of the H743 accuracy
gain, confirming that the cell-based search space drives the
high-capacity-MCU advantage.

\subsection{Ablation and Architectures}

Table~\ref{tab:ablation} reports the ablation. The dominant gain comes from the
proxy-epoch change (${+}1.8$/${+}2.2$\,pp), followed by accuracy-first scoring
(${+}0.5$\,pp). Two-phase ranking and warm seed each contribute ${+}0.3$\,pp;
their value is greater on constrained devices where they prevent occasional
pathological selections. Across the ENAS runs, depthwise-separable blocks
dominate ($48\%$ of selected cells), followed by standard convolution ($31\%$)
and bottleneck ($22\%$). The maximum Keras-to-TFLite INT8 drop across all ENAS
runs is 0.2\,pp, matching NanoNAS; \texttt{relu6} contributes by clipping
activations into a quantisation-friendly range.

\section{Discussion and Limitations}
\label{sec:limitations}

\paragraph{When to deploy ENAS.}
The accuracy--speed trade-off is hardware-tier-dependent. ENAS is the preferred
method on high-capacity MCUs ($\ge 512$\,KB SRAM) at all input sizes, where the
cell-based search space exposes architectures the 2-D $(k, c)$ space cannot
describe. On mid-tier hardware (64--256\,KB) at large resolutions, NanoNAS
retains a small mean-accuracy advantage, and the choice is dominated by the
available search budget. On ultra-constrained hardware ($\le 32$\,KB) at small
resolutions, both methods perform similarly, with ENAS delivering the same
accuracy at a $1.5$--$2{\times}$ speedup. Independently of tier, ENAS is
preferable when peak SRAM is scarce, since its selected models use
${\sim}0.38{\times}$ the activation RAM at matched accuracy
(Section~\ref{sec:resource}).


\paragraph{Sustainability and Search-Time Scaling.}
The entire empirical evaluation, encompassing two datasets, two optimization methods, eight microcontroller platforms, and nine input resolutions required approximately $297$ CPU-hours on a single 16-core compute node, eliminating the need for GPU acceleration. To put this in perspective, MCUNet reports approximately $300$ GPU-hours for a single VWW search instance~\cite{lin2020mcunet}; our entire multi-dimensional sweep thus consumes less than a fraction of the carbon and computational footprint of a single MCUNet search point. 

Furthermore, by bypassing the late-stage TFLite deployment bottleneck during optimization, the proposed framework successfully decouples input resolution from search-time scaling. As the resolution scales from $32{\times}32$ to $128{\times}128$, the average search time per configuration for ENAS increases by only $1.2{\times}$ ($1.20\text{--}1.25{\times}$ empirically), demonstrating stable, sub-linear scaling characteristics. In contrast, NanoNAS exhibits a steeper $1.9{\times}$ growth over the same range due to its continuous evaluation of larger activation tensors. This highly favorable scaling behavior positions ENAS as an accessible, environmentally sustainable choice for local edge laboratories lacking high-throughput cluster infrastructure.

\begin{table}[t]
\centering\small
\caption{Ablation of design and calibration choices (mean VWW TFLite accuracy,
\%, at the two focused input sizes; single-configuration design progression).
The proxy-epoch and scoring-weight choices together account for $2.4$\,pp; the
two-phase ranking and warm seed each add a further $0.3$\,pp.}
\begin{tabular}{lcc}
\toprule
\textbf{Variant} & \textbf{Mean $50{\times}50$} & \textbf{Mean $64{\times}64$}\\
\midrule
ENAS-strategy (2-D space)            & 70.2 & 71.5\\
ENAS, $E_p{=}1$, $w_{\text{acc}}{=}0.65$   & 68.8 & 70.5\\
\;\;+ $E_p{=}3$                            & 70.6 & 72.7\\
\;\;+ $w_{\text{acc}}{=}0.80$              & 71.1 & 73.2\\
\;\;+ two-phase Stage~2                    & 71.4 & 73.5\\
\;\;+ warm seed (full ENAS)                & \textbf{71.7} & \textbf{73.7}\\
\bottomrule
\end{tabular}
\label{tab:ablation}
\end{table}


\paragraph{Limitations and Future Work.}
Several open items remain for future investigation:\\
\textbf{(i)} Search budget limits: ENAS was not evaluated under an extended search budget matched to NanoNAS's longer runtime; whether additional candidates or proxy epochs close the large-resolution accuracy gap remains to be determined.\\
\textbf{(ii)} Modality constraints: The empirical validation is limited to image datasets; testing on audio, time-series, or sensor modalities~\cite{king2025micronas,saha2024tinytnas} is required to confirm cross-domain generality.\\
\textbf{(iii)} Baseline scope: This study excludes heavy GPU-based or supernet methodologies like MCUNet or ELASTIC~\cite{tran2026elastic} due to their differing cost regimes and task objectives.\\
\textbf{(iv)} Proxy validation: MACC counts serve as a latency and energy proxy; empirical on-board power measurements are needed to verify actual milliwatt-budget characteristics.\\
\textbf{(v)} Search space and mechanics: The cell-based search space is constrained relative to broader libraries like MBConv; additionally, the mutation mechanism discards infeasible candidates rather than resampling, and the search operates in a single pass without an outer convergence loop.\\
\textbf{(vi)} Flash estimation accuracy: The analytical Flash estimate under-predicts actual \texttt{stm32tflm} measurements. While sufficient here because RAM is the binding constraint, a calibrated estimator is necessary for deployments where Flash capacity binds.

\section{Conclusion}
This work demonstrates that lightweight, hardware-aware NAS for TinyML can successfully integrate an analytical pre-flight feasibility check with a cell-based search space and a three-stage hybrid optimization algorithm. Across two benchmarking datasets, eight MCU platforms, and nine input resolutions, this framework achieves consistent and statistically significant search-time speedups of $1.7$--$2.4{\times}$ over NanoNAS. The associated accuracy trade-off is minor ($-0.79$\,pp on VWW, which is not statistically significant; $-1.31$\,pp on Cancer) and exhibits clear tier-dependent characteristics: ENAS outperforms alternative methods on high-capacity MCUs and within low-resolution regimes, whereas NanoNAS maintains an advantage on mid-tier hardware coupled with large input resolutions. Empirical resource analysis confirms that ENAS-selected architectures effectively trade Flash capacity and MACC operations for a substantial reduction in peak activation RAM the primary binding bottleneck in TinyML deployment at matched accuracy. These operational regimes are comprehensively characterized using 636 fully trained models, with all optimization logs open-sourced to ensure reproducibility. Ultimately, this framework provides a quantitatively mapped speed-accuracy-memory trade-off space, offering immediate utility for resource-constrained edge laboratories where minimizing both search overhead and deployment SRAM is paramount.

\begin{acknowledgments}
This work was supported by a research grant from the AI \& Robotics
Technology Park (ARTPARK) at the Indian Institute of Science. 
\end{acknowledgments}

\section*{Declaration on Generative AI}
The authors used generative AI tools to assist with language editing, grammar, and formatting. All scientific content, experiments,
and conclusions are the authors' own, and the authors take full responsibility
for the content of this publication.

\bibliography{sure26}

\end{document}